\documentclass[10pt,twocolumn,letterpaper]{article}

\usepackage{cvpr}              

\usepackage[accsupp]{axessibility}
\usepackage{graphicx}
\usepackage{amsmath}
\usepackage{amssymb}
\usepackage{balance}
\usepackage{booktabs}
\usepackage{multicol}
\usepackage{multirow}
\usepackage{subfigure}
\usepackage{colortbl}
\usepackage{enumitem}
\usepackage[normalem]{ulem}
\usepackage[T1]{fontenc}
\usepackage{dsfont}
\usepackage[dvipsnames]{xcolor}
\usepackage[pagebackref=false,breaklinks,colorlinks]{hyperref}

\usepackage[capitalize]{cleveref}
\crefname{section}{Sec.}{Secs.}
\Crefname{section}{Section}{Sections}
\Crefname{table}{Table}{Tables}
\crefname{table}{Tab.}{Tabs.}
\pdfoutput=1
\definecolor{Gray}{gray}{0.9}
\def\cvprPaperID{2355} 
\def\confName{CVPR}
\def\confYear{2022}
\newcommand{\formattedparagraph}[1]{\noindent \textbf{#1}}
\newcommand{\etals}{\text{et al.}}
\newcommand{\ies}{\text{i.e.}}
\begin{document}

\title{Uncertainty-Aware Deep Multi-View Photometric Stereo}

\author{\quad Berk Kaya$^{1}$\quad Suryansh Kumar$^{1}\thanks{Corresponding Author ({\tt\small k.sur46@gmail.com})}$ \quad Carlos Oliveira$^1$ \quad Vittorio Ferrari$^{2}$\quad Luc Van Gool$^{1, 3}$\\
ETH Z\"urich${^1}$, Google Research$^2$, KU Leuven$^3$
}
\maketitle

\begin{abstract}
This paper presents a simple and effective solution to the longstanding classical multi-view photometric stereo (MVPS) problem. It is well-known that photometric stereo (PS) is excellent at recovering high-frequency surface details, whereas multi-view stereo (MVS) can help remove the low-frequency distortion due to PS and retain the global geometry of the shape. This paper proposes an approach that can effectively utilize such complementary strengths of PS and MVS. Our key idea is to combine them suitably while considering the per-pixel uncertainty of their estimates. To this end, we estimate per-pixel surface normals and depth using an uncertainty-aware deep-PS network and deep-MVS network, respectively. Uncertainty modeling helps select reliable surface normal and depth estimates at each pixel which then act as a true representative of the dense surface geometry. At each pixel, our approach either selects or discards deep-PS and deep-MVS network prediction depending on the prediction uncertainty measure. For dense, detailed, and precise inference of the object's surface profile, we propose to learn the implicit neural shape representation via a multilayer perceptron (MLP). Our approach encourages the MLP to converge to a natural zero-level set surface using the confident prediction from deep-PS and deep-MVS networks, providing superior dense surface reconstruction. Extensive experiments on the DiLiGenT-MV benchmark dataset show that our method provides high-quality shape recovery with a much lower memory footprint while outperforming almost all of the existing approaches.
\end{abstract}

\section{Introduction}
In the coming decade, dense 3D data acquisition of objects is likely to become one of the most important problems in computer vision and industrial machine vision. Moreover, it can be helpful for a wide range of other cutting-edge scientific disciplines such as metrology \cite{furukawa2015multi}, geometry processing \cite{bronstein2008numerical}, forensics \cite{moons20093d}, etc.
At present, it is widely accepted that methods such as structure-from-motion \cite{schoenberger2016sfm, hartley2003multiple}, multi-view stereo \cite{furukawa2015multi}, photometric stereo \cite{woodham1980photometric, kaya2020uncalibrated, sarno2022neural}, and other standalone approaches \cite{newcombe2011kinectfusion, wolff1997polarization, mildenhall2020nerf, kumar2019superpixel, kumar2017monocular} are not sufficient on their own to provide detailed and precise 3D reconstruction for all kinds of surfaces \cite{moons20093d}. Therefore, methods that combine complementary surface estimates by leveraging more than one modality are often preferred \cite{nehab2005efficiently, li2020multi}.

\begin{figure*}[t]
    \centering
    \subfigure[\label{fig:example_MVS}{Multi-View Stereo (MVS) Reconstruction}]
    {\includegraphics[width=0.31\linewidth, height=0.19\textwidth]{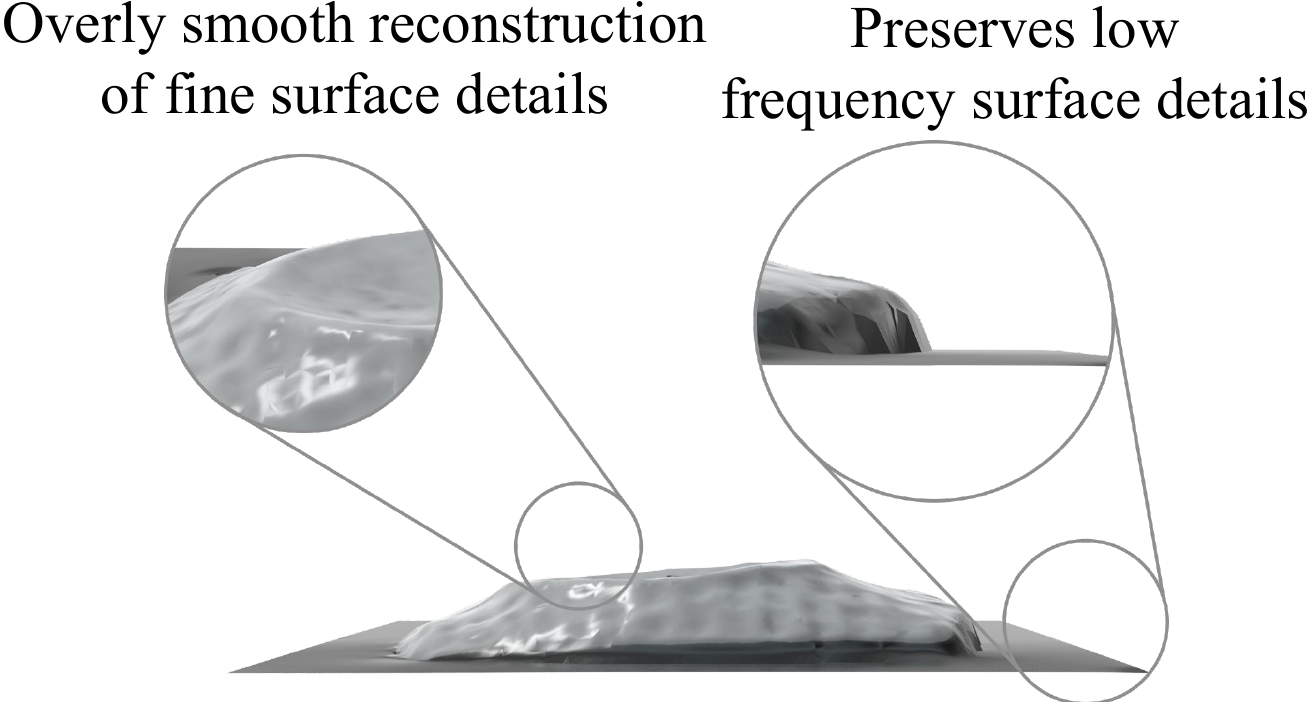}}
    ~~\subfigure[\label{fig:example_PS}{Photometric Stereo (PS) Reconstruction}]
    {\includegraphics[width=0.31\linewidth, height=0.19\textwidth]{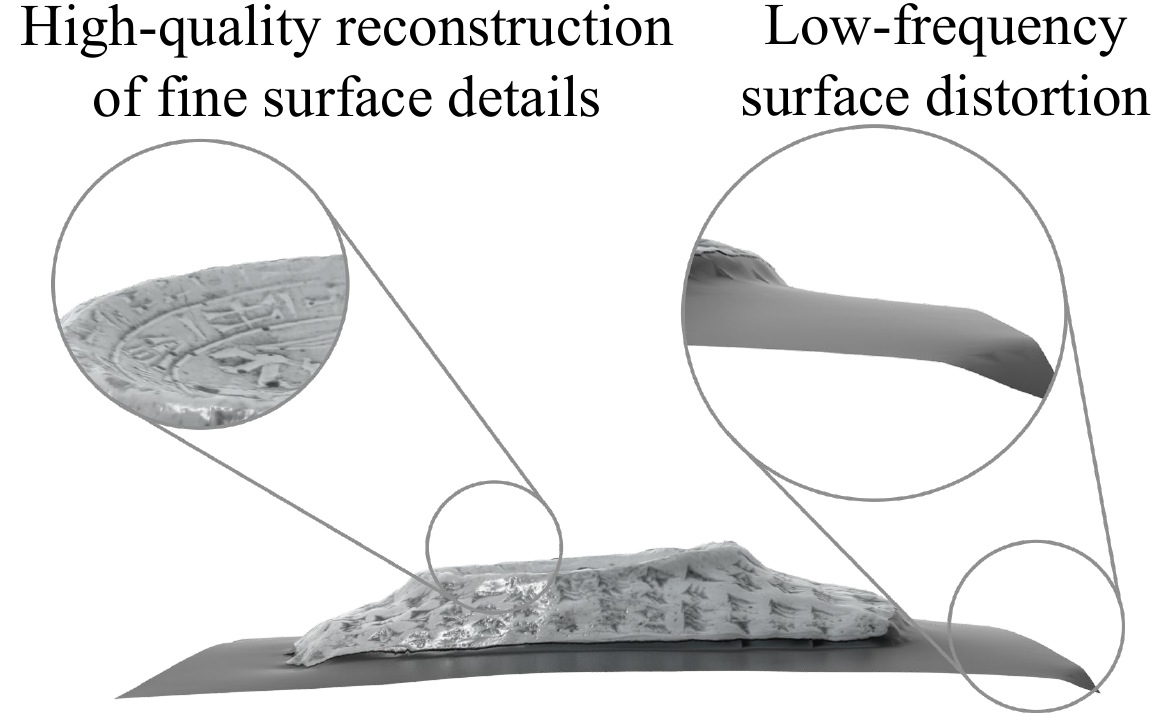}}
    ~~\subfigure[\label{fig:example_fusion}{Our Approach}]
    {\includegraphics[width=0.32\linewidth, height=0.19\textwidth]{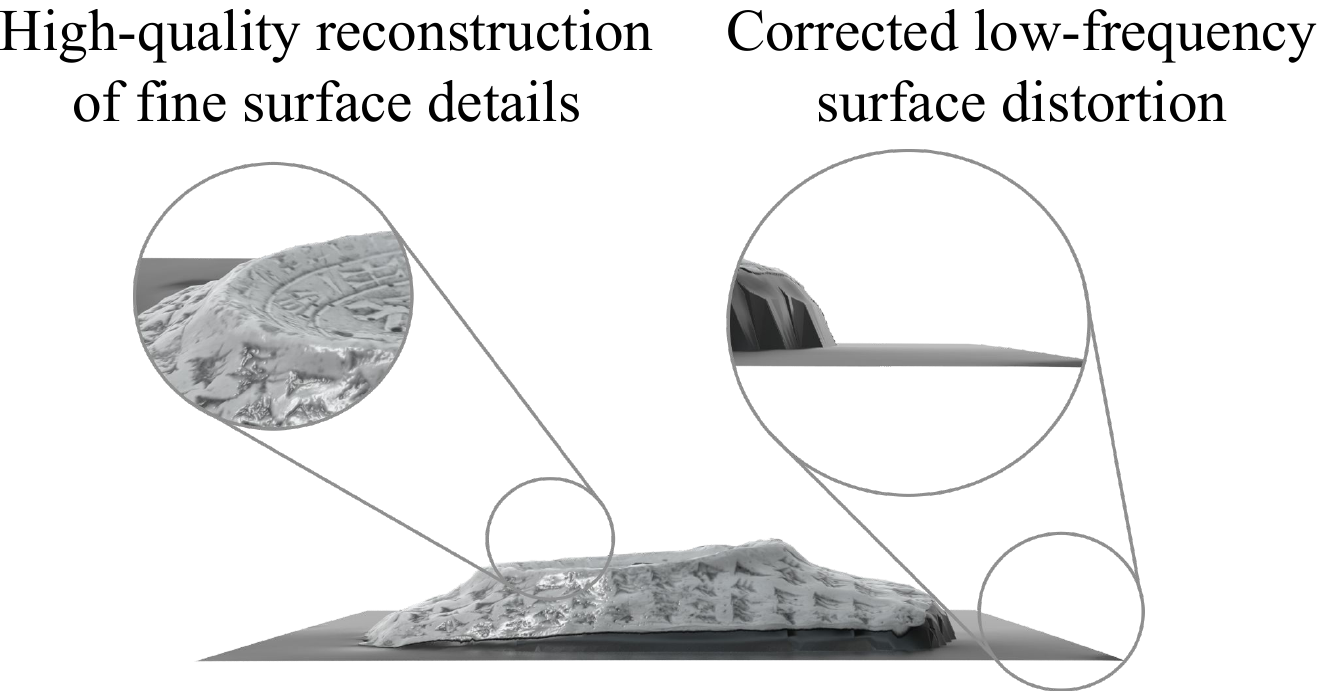}}
    \caption{\footnotesize Our method handles the high and low-frequency surface components quite well. It overcomes the high and low-frequency surface reconstruction problem by suitably utilizing the complementary surface estimates via uncertainty modeling, and neural level set optimization (a) MVS reconstruction preserves the plane geometry but loses the finer details. (b) PS captures the fine geometric details but introduces global distortion. (c) Our approach. }
    \label{fig:illustration}
\end{figure*}

Among the passive 3D shape acquisition methods, multi-view stereo (MVS) has become the most popular approach \cite{wu2011visualsfm, schoenberger2016sfm, furukawa2009accurate}, especially after the proliferation of cheap digital cameras for high-quality imaging. Yet, MVS works best for Lambertian textured surfaces and gives unreliable results for non-textured objects with non-Lambertian surface reflectance property. Moreover, high-frequency surface details such as indentations and scratches are difficult to recover using MVS methods (see Fig.\ref{fig:example_MVS}).

On the other hand, photometric stereo (PS) is magnificent at recovering high-frequency surface details using light-varying images \cite{woodham1980photometric}. It is also effective for non-textured, and non-Lambertian surfaces \cite{chen2019self}. PS allows the recovery of per-pixel depth of the object by integration of the estimated surface normals \cite{horn1986variational}. However, PS suffers from the main shortcoming: The recovered surface profile is globally deformed by a low-frequency distortion \cite{nehab2005efficiently}.  Such distortion is likely due to numerical integration of the surface normal map without explicit constraints between multiple disconnected regions of the object's surface \cite{nehab2005efficiently, xie2019surface, queau2018variational} or non-directional lighting effects (see Fig.\ref{fig:example_PS}).

Combining the complementary responses of MVS and PS is known as multi-view photometric stereo (MVPS) \cite{hernandez2008multiview}. In this paper, we propose an approach that can effectively exploit such complementary surface information. Our work leverages recent advances in deep neural networks. To this end, we use a PatchMatch-based deep-MVS network \cite{wang2021patchmatchnet} to infer the per-pixel depth and a CNN-based deep-PS network \cite{ikehata2018cnn} to infer per-pixel surface normal. But, we know that such deep network models have their accuracy limits and can predict erroneous depth or surface normals in certain parts of the object. In that case, if we naively combine the output predicted by these networks, we may end up having a bad overall result. To resolve this, we extend the deep-PS and deep-MVS networks with per-pixel uncertainty estimation capability. Using the prediction uncertainty as a measure, we select and combine only reliable surface estimates at each pixel, that is either deep-PS normal, deep-MVS depth, both, or none of the prediction.

Using our approach of selecting and discarding surface estimates may result in the loss of some pixels' corresponding 3D surface details. To recover those lost geometric details, we introduce a neural network (MLP layers only) based optimization to recover the overall dense shape from those selected surface predictions by representing the object's shape as level sets of the neural network. Our overall loss function optimization encourages the zero-level set of the neural network to converge to the confident surface estimates. To that end, we first convert the depth estimate to point cloud while keeping the predicted surface normals representation as it is. Our approach then optimizes for parameters of an MLP so that it approximates a signed-distance-function (SDF) to a plausible surface based on the point cloud, surface normals, and an implicit geometric regularization term developed on the Eikonal partial differential equation \cite{crandall1983viscosity}. Fig.\ref{fig:example_fusion} shows an example reconstruction using our approach. In summary, we make the following contributions:

\begin{itemize}[leftmargin=*,topsep=0pt, noitemsep]
    \item We present an effective and easy-to-use deep neural network-based solution to the classical MVPS problem for dense, detailed, and precise recovery of 3D shapes.
    
    \item We introduce uncertainty-aware deep-PS and deep-MVS modeling in the MVPS pipeline. Modeling uncertainty helps as a measure in automatic discarding of unreliable surface estimates at a pixel, hence improving robustness.
    
    \item We propose an implicit neural shape representation based on the Eikonal term in the MLP loss for natural zero level set surface recovery \cite{crandall1983viscosity}. It reliably infers the object's dense surface geometry defined by the confident deep-PS and deep-MVS prediction with better memory foot-print than the methods based on mesh processing \cite{li2020multi, park2016robust}.
\end{itemize}

\section{Related Work}
To better place the previous works, we have divided the important MVPS methods literature into two groups.

\formattedparagraph{(a) Traditional MVPS Methods.}
MVPS is a classical 3D shape acquisition setup introduced by Hernandez \etals \cite{hernandez2008multiview}. The proposed setup is composed of a turn-table with an object placed at the center of the table for multi-view and photometric stereo image acquisition. The MVPS algorithm proposed by Hernandez \etals \cite{hernandez2008multiview} combines the multi-view PS results and corrects its low-frequency surface distortion via multi-view geometric constraint. Yet, the method works well only for specific parametric BRDF models \cite{li2020multi}. Later, Park \etals \cite{park2013multiview, park2016robust} proposed an uncalibrated MVPS method to recover fine geometric details of the shape. It requires an initial coarse mesh with a 2D displacement map for optimization leading to shape recovery.
Still, the method cannot handle objects with diverse surface reflectance properties. Further, it often fails on textureless regions with non-Lambertian reflectances \cite{li2020multi}. Logothetis \etals \cite{logothetis2019differential} used a volumetric approach to solve MVPS via a variational framework. Recently, Li \etals \cite{li2020multi} proposed a systematic geometric approach to MVPS showing state-of-the-art (SOTA) results. However, it consists of several carefully crafted explicit geometric modeling steps such as iso-depth contour estimation, tracing contours, multi-view depth propagation, point sorting, shape optimization, etc.  It requires a successful execution of each of these steps applied in a sequel to recover the surface. Hence, re-implementing such an approach is complex, strained, and time-consuming.  Furthermore, the use of classical PS and MVS in their pipeline has its limitations; for e.g., classical PS, MVS may not handle a wide range of objects with non-Lambertian properties, etc.  

On the contrary, we propose a novel, simple, and everyday deep neural network based solution to MVPS that can handle objects with different reflectance properties and delivers 3D reconstruction accuracy as good as the complex SOTA when tested on the benchmark dataset \cite{li2020multi}.

\formattedparagraph{(b) Deep Learning based MVPS Methods.}
Kaya \etals \cite{kaya2021neural} recently proposed a neural radiance fields-based approach to solve MVPS. It uses a pre-trained deep-PS model to predict the surface normal. It then conditions the multi-view image rendering on the predicted PS surface normal to recover the surface geometry. Even though the method is simple and usable, the recovered geometry is poor in quality. Our literature review shows a lack of a robust modern neural network approach to solve MVPS, which is excellent at learning the object's surface properties from data. Thus, it has become increasingly evident that a simple and effective learning method is essential for the MVPS problem. 

Other related work uses an active 3D sensing modality with PS. For instance, Nehab \etals \cite{nehab2005efficiently} used a structured lighted scanner, whereas Chatterjee \etals \cite{chatterjee2015photometric} relied on a RGB-D sensor to measure the 3D position data. Instead, our work focuses on the classical MVPS setup \cite{hernandez2008multiview, park2013multiview, li2020multi}. It has some apparent advantages over active 3D scanning methods. Firstly, it is easy and cost-effective to perform high-quality image acquisition, as regular cameras are sufficient. Secondly, it is relatively noise-free and gives dense per-pixel information compared to incomplete range data with outliers provided by structured light \cite{geng2011structured}, 3D laser scanner \cite{davis2003spacetime}, and depth sensors \cite{zollhofer2018state}.

\section{Proposed Approach}
We denote $\mathcal{I}^{v} = \{I_1^{v},\dots,I_{N_p}^{v}\}$ as the set of $N_p$ input PS images for a given view\footnote{In a turn-table setup, each view captures different part of the object.} $v \in \{1,\dots,N_m\}$, where $N_m$ is the number of views. For MVS, we follow Li \etals~\cite{li2020multi} work, which take the median of all PS images per camera view to have MVS images.
Concretely, $Y^{v} = \textrm{median}(\mathcal{I}^v)$ gives us the set $\mathcal{Y} = \{Y^1,\dots,Y^{v},\dots,Y^{N_m}\}$ of MVS images. We assume a calibrated setting, i.e., light source directions, intensities, and camera calibrations are known. 

The rest of the section is organized as follows: 
In Sec. \S \ref{subsec:ps_network}, we introduce our uncertainty-aware deep-PS network. Next, in Sec. \S \ref{subsec:mvs_network}, we describe the uncertainty-aware deep-MVS network pipeline. Finally, in Sec. \S \ref{subsec:implicit_neural_representation},  we explain our neural shape representation approach and level set optimization for dense, detailed, and accurate 3d surface recovery from high-fidelity surface normals and depth estimates.

\subsection{Uncertainty-Aware Deep Photometric Stereo} \label{subsec:ps_network}
When an object surface is illuminated by $k^{th}$ point light source located in the direction $l_k \in \mathbb{R}^{3 \times 1}$, then the PS image $I_{k}^{v} \in \mathbb{R}^{p \times 1}$ (vectorized form) captured by a camera in the view direction $\mathbf{v} \in \mathbb{R}^{3 \times 1}$ w.r.t the object is modeled as:
\begin{equation}
    \centering
    \begin{aligned}
    I_{k}^{v} = e_{k} \cdot \Psi(\mathbf{N}_v, l_k,  \mathbf{v}) \cdot \max(\mathbf{N}_{v}^{T}l_k, 0)  + \epsilon_k  .
    \end{aligned}\label{eq:image_formation_model_ps}
\end{equation}
Here, $p$ symbolises the total number of object surface pixels, $e_k \in \mathbb{R}_+$ denotes the light intensity, $\epsilon_k$ is an additive error.
The image formation model in Eq. \eqref{eq:image_formation_model_ps} assumes a real-world object with a reflective surface, whose appearance is encoded by a BRDF $\Psi$ with surface normal $\mathbf{N}_v \in \mathbb{R}^{ 3 \times p}$ and $\max(\mathbf{N}_v^{T}l_k, 0)$ accounts for the attached shadows. The formulation in Eq.\eqref{eq:image_formation_model_ps} has led to outstanding developments in PS for recovering fine surface details \cite{woodham1980photometric, wu2010robust, shi2016benchmark}. Yet, modeling unknown reflectance of different objects remains a challenge. Recently, deep learning-based methods have been proposed to utilize neural networks' ability to learn complicated BRDFs from input data \cite{santo2017deep, shi2019tpami, chen2019self, taniai2018neural, ikehata2018cnn, kaya2020uncalibrated}.
Accordingly, we adhere to using a supervised deep learning framework to have high-fidelity surface normal predictions at test time for a diverse set of objects with different material properties. To that end, we adopt an observation map based modeling in deep-PS \cite{ikehata2018cnn} due to its simplicity and notable performance on PS benchmark datasets \cite{shi2016benchmark, shi2019tpami}.

\begin{figure*}[t]
    \centering
    \includegraphics[width=0.96\textwidth]{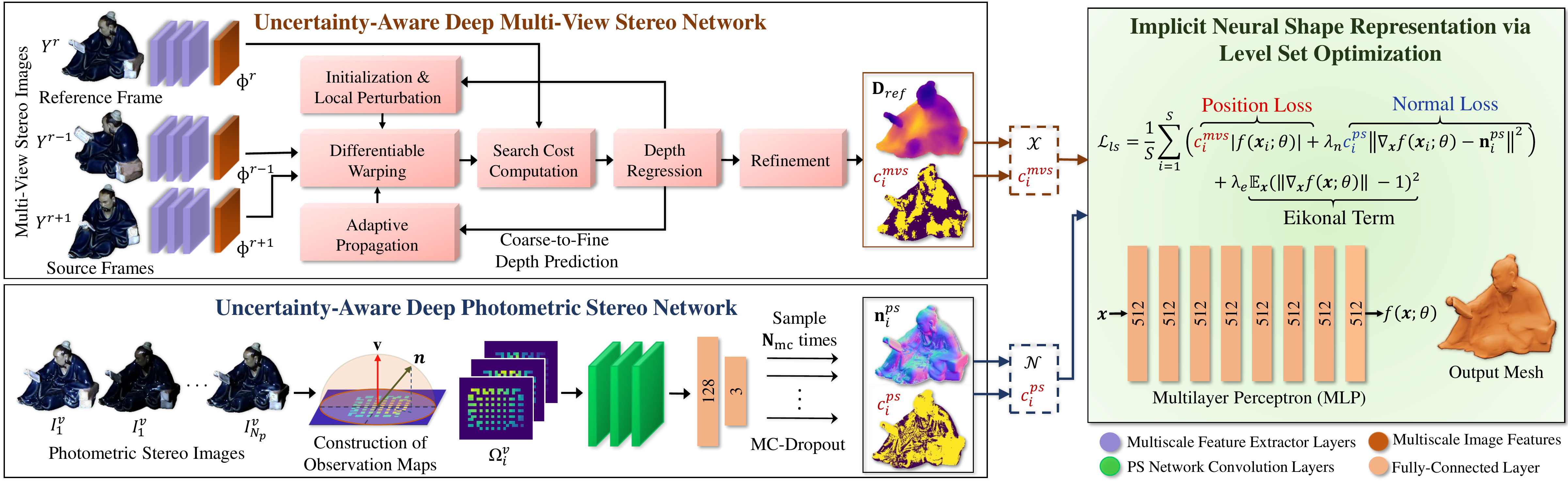}
    \caption{\footnotesize
     An overview of our uncertainty-aware deep-MVPS approach.  We first predict per-pixel depth and surface normals via deep-MVS and deep-PS networks. Then, we recover dense, detailed 3D shape by using these surface estimates in neural level set optimization.
     Our approach utilizes prediction uncertainty as a measure of reliability and uses highly confident surface estimates in optimization for better surface recovery (Best View on Screen).
    }\label{fig:pipeline}
\end{figure*}

\smallskip
\noindent
\textbf{$\bullet${~Observation Map.}}
An observation map is a 2D matrix that stores normalized grayscale pixel values observed under different light sources (Fig.\ref{fig:pipeline} bottom left). We independently create observation map for every object pixel. In a PS setup, there exists a one-to-one mapping between $k^{th}$ light source direction $[l_{k}(x), l_{k}(y), l_{k}(z)]^{T} \in \mathbb{R}^{3 \times 1}$ and its corresponding $x-y$ coordinate projection $[l_{k}(x), l_k(y)]^{T} \in \mathbb{R}^{2 \times 1}$. Utilizing these notations, observation map $\Omega_{i}^{v} \in \mathbb{R}^{\omega \times \omega}$ for $i^{th}$ pixel captured from view $v$ is defined as:
\begin{equation}  \label{eq:observation_map}
    \Omega_{i}^{v}\Big(\zeta\big(\omega\cdot \frac{(l_k(x) + 1)}{2} \big), \zeta\big(\omega\cdot \frac{(l_k(y) + 1)}{2} \big) \Big) =  \frac{I_{k}^{v}(i)}{\eta_{i} e_k}
\end{equation}
where, $\omega$ is the 2D resolution of the observation map and the scalar $\eta_i = \max(I_1^{v}(i)/e_1,\dots, I_{{N}_{p}}^{v}(i)/e_{{N}_{p}})$ is the normalizing constant. Here, $\zeta: \mathbb{R} \mapsto \mathbb{Z}_{0}^+$, since the projected source direction vectors can take values from $[-1, 1]$, they are scaled to suitable positive integer (including zero).

\noindent
\textbf{$\bullet${~Deep-PS Network Architecture.}}
The deep-PS network takes $\Omega_{i}^{v}$ of each pixel as input and regresses to corresponding ground-truth surface normal at train time to obtain deep-PS model (see Fig.\ref{fig:pipeline}). The network design comprises an initial $3 \times 3$ convolution layer, which converts the observation map into a feature block. Next, the architecture has two dense blocks with a transition layer in between. The dense block is composed of a ReLU and $3 \times 3$ convolution layer. The transition layer comprises a ReLU, $1 \times 1$ convolution, and an average pooling layer. After the second dense block, one convolution and two fully connected layers are applied, followed by a normalization operation to recover unit length surface normals as output.  For more details refer  \cite{ikehata2018cnn}.

\noindent
\textbf{$\bullet${~Uncertainty Modeling in Deep-PS Network.}}
The above deep-PS network is not apt for estimating the uncertainty of the predicted surface normals at test time. For our problem, it is imperative to have that information as perfect prediction is not always possible. Accordingly, we modify the deep-PS architecture to provide uncertainty of the predicted surface normals by leveraging the Bayesian neural network (NN) approach \cite{mackay1992practical, neal2012bayesian}.
Generally, Bayesian NNs are a simple extension of NNs by placing a prior distribution (generally Gaussian) over the NN's weights. Compared to standard NN, it has the advantage of providing an uncertainty measure of the network prediction \cite{polson2017deep, gawlikowski2021survey}. For completeness, let's briefly review the standard Bayesian NN framework.

Let $\{ \mathbf{A}, \mathbf{B} \}$ be the training dataset with $\mathbf{A}$, $\mathbf{B}$ as input and output sets, respectively.
Assume, a Bayesian NN with $L$ layers parameterized by weight $\mathbf{w} = \{\mathbf{W}_j\}_{j=1}^{L}$ with $\mathbf{W}_j$ as the weight matrix for layer $j$.
The predictive distribution $P(\mathbf{b}^{*}|\mathbf{a}^{*}, \mathbf{A}, \mathbf{B})$ for a new input $\mathbf{a}^{*}$ is formulated as $  P(\mathbf{b}^{*}|\mathbf{a}^{*}, \mathbf{A}, \mathbf{B}) = \int P(\mathbf{b}^{*}|\mathbf{a}^{*}, \mathbf{w}) P(\mathbf{w}|\mathbf{A}, \mathbf{B})d\mathbf{w}$. However, to determine predictive distribution value is intractable as $P(\mathbf{w}|\mathbf{A}, \mathbf{B})$ is hard to solve analytically \cite{gal2016dropout}. Generally, variational inference $(VI)$ is used to approximate $P(\mathbf{w}|\mathbf{A}, \mathbf{B})$. By introducing the variational distribution ${Q}_{\gamma}(\mathbf{w})$ parameterized by $\gamma$, the KL divergence between $Q_{\gamma}(\mathbf{w})$ and $P(\mathbf{w}|\mathbf{A}, \mathbf{B})$ is minimized as $\mathcal{L}_{VI} = - \int  {Q}_{\gamma}(\mathbf{w}) log(\mathbf{B} | \mathbf{A},  \mathbf{w})d\mathbf{w} + KL( {Q}_{\gamma}(\mathbf{w}) || P(\mathbf{w}))$.

\smallskip
While $VI$ based on KL divergence is widely used, the assumption of Gaussian distribution on the network parameters increases the complexity of the model. That, in turn, reduces the model efficiency without a significant gain in predictive power. 
Not long ago, Gal \etals \cite{gal2015bayesian} proposed a Bernoulli distribution-based $VI$ approach which provides a simpler and fruitful approximation to the posterior distribution. To adopt Bernoulli distribution approach to our deep-PS network, we model $\mathbf{W}_j = \mathbf{M}_j \cdot \textrm{diag}([z_{j,u}]_{u=1}^{K_{j}})$ and $z_{j,u}$ as $\text{Bernoulli}(q_j) ~\text{for} ~j = 1,\dots, L, ~u=1,\dots, K_{j-1}$. Here, $\mathbf{M}_j$ denotes variational parameters of $\mathbf{W}_j \in \mathbb{R}^{K_{j} \times K_{j-1}}$ with $K_{j}$ the number of units at layer $j$, and $z_{j,u}$ denotes the Bernoulli random variables with probability $q_j$.

With such parameterization, the integral term in $\mathcal{L}_{VI}$ is approximated by sampling $\mathbf{W}$ from a Bernoulli distribution, also known as Monte Carlo (MC) integration approach. Likewise, the KL divergence term in $\mathcal{L}_{VI}$ is replaced with a weight decay on network parameters \cite{gal2015bayesian}. Using these approximations, we train our uncertainty-aware deep-PS network using the following loss function:
\begin{equation}\label{eq:ps_network_loss}
  \mathcal{L}_{ps} = \frac{1}{{N}_{mc}} \sum_{j=1}^{{N}_{mc}} || \mathbf{\tilde{n}}_j - \mathbf{n}_{gt} ||_2^2 + \lambda_{w} \sum_{j=1}^{L} || \mathbf{W}_j ||_2^2
\end{equation}
where, ${N}_{mc}$ is the number of MC samples, $\mathbf{\tilde{n}}_j$ is the estimated surface normal in each forward pass and $\mathbf{n}_{gt}$ is the ground-truth surface normal. As shown in Gal \etals \cite{gal2015bayesian} work, we can realize the approximation to Bernoulli distribution by introducing dropout layers in the neural network. Accordingly, we apply dropout with $q_j = 0.2$ after each convolution and fully-connected layer in deep-PS network. We keep dropout layers active during train and test time.

Due to the introduction of dropout layers, we now have a stochastic network. At test time, we run the trained model multiple times, recording the (potentially varying) surface normal prediction at each $i^{th}$ pixel. We then calculate the mean and variance of these multiple predictions at every pixel. The mean is taken as the final prediction $\mathbf{n}_{i}^{ps} \in  \mathbb{R}^{3 \times 1}$ and the variance as its uncertainty $\sigma_{i}^{2} \in \mathbb{R}^{3 \times 1}$.
Since our approach is focused on highly confident predictions, we convert the per-pixel variance to a single binary variable $c_{i}^{ps}$. We set $c_{i}^{ps} = 1$, if $\| \sigma_i^{2}\|_{1} < \tau_{ps}$ and $c_{i}^{ps} = 0$, otherwise.

\subsection{Uncertainty-Aware Deep Multi-View Stereo} \label{subsec:mvs_network}
Similar to deep-PS network, we aim to have a per-pixel uncertainty measure but now on the depth prediction. One natural way is to similarly use MC dropout strategy to deep-MVS network. Fortunately, there already exist deep MVS frameworks which have the intrinsic ability to implicitly provide uncertainty measure via confidence values of their depth predictions \cite{yao2018mvsnet, wang2021patchmatchnet, chen2019point, huang2018deepmvs}. Hence, it is inefficient to add extra complexity by introducing MC dropout layers. Among \cite{yao2018mvsnet, wang2021patchmatchnet, chen2019point, huang2018deepmvs}, we use \cite{wang2021patchmatchnet}, \ies, PatchMatch based deep-MVS network due to its recent SOTA performance on MVS benchmarks and fast inference on large scale images.

\noindent
\textbf{$\bullet${~PatchMatch based Deep-MVS Network.}} Similar to PatchMatch algorithm \cite{barnes2009patchmatch}, the PatchMatch based deep-MVS network employs \cite{barnes2009patchmatch} via similar three steps (but in 3d scene space) as follows: \textit{\textbf{(i)}} Initialization step: Generating depth hypotheses, \textit{\textbf{(ii)}} Propagation step: Propagate the hypotheses to neighbors, and \textit{\textbf{(iii)}} Evaluation step: Compute the similarity cost and search for best solution. We apply these steps on per-pixel multi-scale features that are hierarchically extracted  from $\mathcal{Y}$ at $M$  different resolution scales \cite{lin2017feature, wang2021patchmatchnet} (see Fig.\ref{fig:pipeline} top left). This allowed us to estimate depth in a coarse-to-fine manner. Before we introduce the steps, let's denote the reference frame by ${Y}^{r} \in \mathbb{R}^{w \times h}$, coordinates of the $i^{th}$ pixel by $\mathbf{y}_i$, frame $r$ feature by $\Phi^{r}$, and camera $r$ intrinsic calibration matrix by $\mathbf{K}_{r}$.
For each reference frame, we pick $N_s$ source frames where ${Y}^{s} \in \mathbb{R}^{w \times h}$ denotes a source frame. 
$(\mathbf{R}_{r, s}, \mathbf{t}_{r, s})$  denotes the relative motion between frame $r$ and $s$. We skip to add extra notation for stage number for simplicity of writing. 

\noindent
\textit{\uline{\textbf{(i)} Initialization.}} In the first iteration, per-pixel $\mathcal{D}_{f}$ depth hypotheses are sampled. Once initialized, ``local perturbations'' are invoked in the subsequent iteration at each stage to diversify the hypotheses \cite{bleyer2011patchmatch}. For local perturbations, per-pixel $N_{l}^m$ hypotheses are generated at stage $m$. 

\noindent
\textit{\uline{\textbf{(ii)} Propagation.}}
Existing hypotheses are enriched using spatially neighboring pixels that are likely to have similar depth values. For that, an ``adaptive propagation'' approach that uses a learnable offset to gather hypotheses from the same physical surface rather than a fixed set of neighbors is used. Accordingly, 2D CNN is applied on $\Phi^{r}$ to learn a 2D offset for each pixel and obtain $N_d^m$ additional hypotheses at stage $m$ using the depth map of the previous iteration.

\noindent
\textit{\uline{\textbf{(iii)} Evaluation.}} The best solution is searched by evaluating the similarity cost for existing depth hypotheses.  For that, group-wise correlation between $\Phi^r(\mathbf{y}_i)$ and warped source feature $\Phi^{s}(\mathbf{y}_{i}^{s, j})$ is calculated for every pixel and depth hypothesis   \cite{xu2020learning}.
Here, group-wise correlation is obtained by dividing the features into groups along channel dimension and computing the inner product per group. $\mathbf{y}_{i}^{s,j}$~is the warped coordinates of the $i^{th}$ pixel at source view $s$ and is computed with ``differentiable warping'' relation $\mathbf{y}_{i}^{s,j} = \mathbf{K}_{s} \Big(\mathbf{R}_{r, s}\big(d_{j}(\mathbf{y}_i) \cdot \mathbf{K}_{r}^{-1} \mathbf{y}_i\big) + \mathbf{t}_{r, s} \Big)$ where $d_j(\mathbf{y}_{i})$ stands for the $j^{th}$ depth hypothesis at pixel coordinates $\mathbf{y}_{i}$.

The group-wise correlation costs over number of views are aggregated with per-pixel view weight \cite{xu2020pvsnet, schoenberger2016sfm}. Then, a 3D convolution layer with $1\times1\times 1$ kernel is applied on the aggregated cost to obtain ``search cost'' per-pixel and depth hypothesis $\mathbf{J} \in \mathbb{R}^{w \times h \times \mathcal{D}}$. The search cost is further aggregated over a spatial window into $\mathbf{\Tilde{J}}(\mathbf{y}_i, j)$ for increased robustness. For per-pixel ``depth regression'', softmax ($\sigma$) is applied to $\mathbf{\Tilde{J}}(\mathbf{y}_i, j)$ 
and expectation over depth hypotheses is evaluated as $    \mathbf{D}(\mathbf{y}_i) = \sum_{j=0}^{\mathcal{D}-1} d_j(\mathbf{y}_{i}) \cdot \sigma( \mathbf{\Tilde{J}}(\mathbf{y}_i, j))$.

Subsequently, per-pixel confidence measure $\rho_i$ is computed using the predicted probability of the most likely depth hypothesis, \ies $\rho_i = \sigma( \mathbf{\Tilde{J}}(\mathbf{y}_i, j^{*}))$.
Finally, an independent depth residual network based on \cite{hui2016depth} is used for ``refinement'' to obtain output depth map $\mathbf{D}_{ref}$. For more details on PatchMatch based deep-MVS network, refer \cite{wang2021patchmatchnet}.

\noindent
\textbf{$\bullet${~MVS Loss Function.}}
We use $l_1$ loss between the estimated depth and the ground-truth depth at the same resolution.  The total MVS loss takes into account the PatchMatch loss $\mathcal{L}_{pm}$ at each stage along with refined depth loss $\mathcal{L}_{ref}$.
\begin{equation}
    \mathcal{L}_{mvs} = \mathcal{L}_{pm} + \mathcal{L}_{ref}, ~\text{where} ~~\mathcal{L}_{pm} = \sum_{m=1}^{M} \sum_{t=1}^{N_\text{iter}^{m}} \mathcal{L}_{t}^{m}
\end{equation}

\noindent
Here, $N_\text{iter}^{m}$ denotes the total number of iterations at stage $m$.

\noindent
\textbf{$\bullet${~Uncertainty Modeling in Deep-MVS Network.}}
To have the notion of per-pixel depth uncertainty, we convert depth prediction confidence value ($\rho_i$'s) to a binary variable $c_{i}^{mvs}$. We set $c_{i}^{mvs} = 1 $ if $\rho_i > \tau_{mvs}$ and $c_{i}^{mvs} = 0$, otherwise.

\begin{figure}
\centering
\subfigure[\label{fig:point_cloud} 2D points ]{\includegraphics[width=0.28\columnwidth]{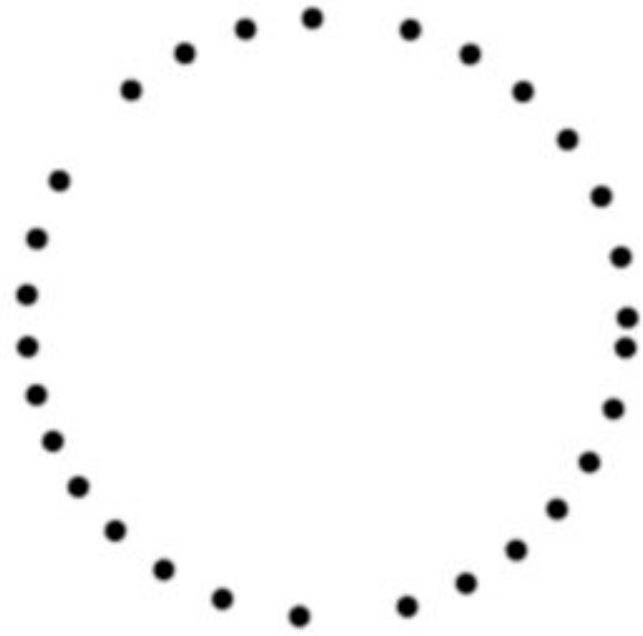}}
~~~~~~~~~~~~~~\subfigure[\label{fig:level_set} MLP level sets]{\includegraphics[width=0.28\columnwidth]{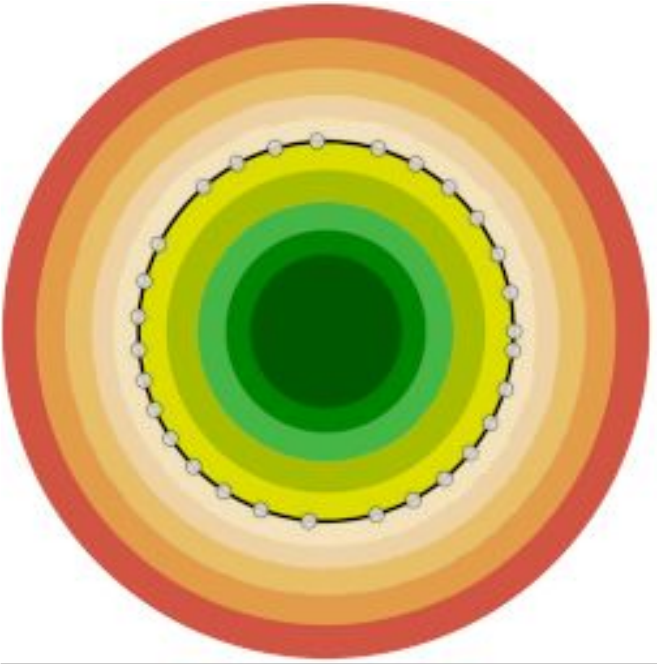}}
\caption{\footnotesize  (a) Sparse 2D point cloud. (b) The Eikonal term based implicit geometric regularization in the optimization gives plausible zero level set for (a) ---shown in black. The different colors in (b) show the level sets. }\label{fig:level_set_illustration}
\end{figure}

\subsection{Implicit Neural Shape Representation} \label{subsec:implicit_neural_representation}
Once we have the complementary information of the object shape, \ies, surface normals, depth, and the fidelity of prediction at hand, our goal is to combine them effectively for dense surface reconstruction. It is pretty natural to go for regular volumetric fusion (VF) approaches \cite{curless1996volumetric, dai2017shape, jiang2020sdfdiff, michalkiewicz2019deep} since surface normals can also provide depth by simple integration \cite{horn1989shape}. However, we know VF uses a fixed size cubic grid independent of the object's geometry, and therefore, may not obey the geometry of the shapes we want to model. Consequently, we propose to work directly on the confident raw surface estimates. For that, we convert deep-MVS network per-pixel depth prediction to point cloud $\mathcal{X} = \{ \mathbf{x}_{i}\}_{i=1}^{{S}} \subset \mathbb{R}^{3}$, where $S$ is the total number object pixel across all views. Additionally, we use per-pixel surface normal prediction from deep-PS network $\mathcal{N} = \{ \mathbf{n}_{i}^{ps}\}_{i=1}^{{S}} \subset \mathbb{R}^{3}$. We propose to recover the 3d surface by optimizing the parameters of a MLP $f(\mathbf{x}; \theta)$. The suggested MLP approximates the signed distance function (SDF) defined by $\mathcal{X}$ and $\mathcal{N}$. We consider the following loss function for MLP optimization:
\begin{equation} \label{eq:level_set_optimization}
    \begin{aligned}
     \mathcal{L}_{ls} & = \frac{1}{S}\sum_{i=1}^{S} \Big(\overbrace{\textcolor{red!80!black}{c_{i}^{mvs}}|f(\mathbf{x}_{i};\theta)|}^{\textrm{\textcolor{red!80!black}{position loss}}}  + \lambda_{n} \overbrace{\textcolor{blue!80!black}{c_{i}^{ps}} ||\nabla_{\mathbf{x}}f(\mathbf{x}_{i}; \theta) - \mathbf{n}_{i}^{ps} ||}^{\textrm{\textcolor{blue!80!black}{normal loss}}}\Big)\\
     & + \lambda_{e} \overbrace{\mathbb{E}_{\mathbf{x}}  ( ||\nabla_{\mathbf{x}}f(\mathbf{x}; \theta)|| -1 )^2}^{\textrm{Eikonal term}}
    \end{aligned}
\end{equation}
The first term encourages the zero level set to converge to high-fidelity position estimates (\ies, for $c_{i}^{{mvs}}=1$). The second term forces the local gradients to be consistent with the reliable normal estimates (\ies, for $c_{i}^{{ps}}=1$). The final term stands for the Eikonal regularization, and it is computed by taking the expectation $\mathbb{E}$ over probability distribution $\mathbf{x} \sim \mathcal{P}$. It is noted over several experiments that the Eikonal term is impressive at implicitly regularizing the zero level set. For more details, refer \cite{gropp2020implicit, crandall1983viscosity}.

\section{Experiment and Results} \label{sec:experiments}

\formattedparagraph{Train set.}
We use CyclesPS synthetic dataset \cite{ikehata2018cnn} to train our \emph{deep-PS network}. It consists of 15 shapes each of which is rendered with diffuse, specular, and metallic BRDF's using 1300 light sources. We use 90\% of the data for training and 10\% for validation. For training the \emph{deep-MVS network}, we use DTU MVS dataset \cite{aanaes2016large}. It provides images of 80 scenes captured from 49 or 64 views (depending on the subject) with their ground-truth (GT) depth maps. We keep the training and validation splits same as outlined in \cite{ji2017surfacenet}. 

\formattedparagraph{Test set.}
We used DiLiGenT-MV benchmark dataset \cite{li2020multi} as the test set to perform all our experiments, statistical evaluations, and ablations. It comprises real-world objects with complex surface profiles and BRDF properties, making it an ideal choice for MVPS algorithm evaluation. It contains MVS and PS images of five real objects captured from 20 viewpoints using the classical turntable MVPS acquisition setup \cite{hernandez2008multiview}. For each view, 96 images are acquired, with each image illuminated by a distinct light source. The distance between the object and the camera center is set to $\sim 1.5m$. In addition, the dataset supplies the light source and camera calibration information. 

\noindent
\textbf{$\bullet${~Implementation Details.}} 
We implemented our approach in Python 3.8 using PyTorch 1.7.1 library. We conducted all the experiments on a commodity desktop supported with NVIDIA GPU with 11GB of RAM. We trained deep-PS and deep MVS networks independently in a supervised setting.

\noindent
\textit{\uline{(i) Uncertainty-Aware Deep Photometric Stereo.}} 
We first generate pixel-wise observation maps $\Omega_i^{v} \in \mathbb{R}^{32\times32}$ using CyclesPS images to train our deep-PS network. For each observation map, we randomly pick between 50 to 1300 light sources. We train the network for 10 epochs using Adam optimizer \cite{kingma2014adam} with a learning rate of $0.1$. During training, we set $\mathbf{N}_{mc} = 10$ and $\lambda_{w}= 10^{-4}$ in our loss function (see Eq.\eqref{eq:ps_network_loss}). After training, we perform uncertainty based inference on DiLiGenT-MV images. For that, we first generate observation maps for each pixel. Then, we run the deep-PS network on each observation map 100 times following MC-Dropout approach \cite{gal2015bayesian}. We calculate, for each pixel, the mean and variance of the outputs to obtain surface normal $\mathbf{n}_{i}^{ps}$ and its uncertainty $\sigma_i^2$. 
We set $\tau_{ps} = 0.03$ to obtain $c_{i}^{ps}$ ($c_{i}^{ps} = 1$ if $\| \sigma_i^{2}\|_{1} < \tau_{ps}$ and $c_{i}^{ps} = 0$, otherwise).

\begin{figure}[t]
    \centering
    \includegraphics[{width=1.0\linewidth}]{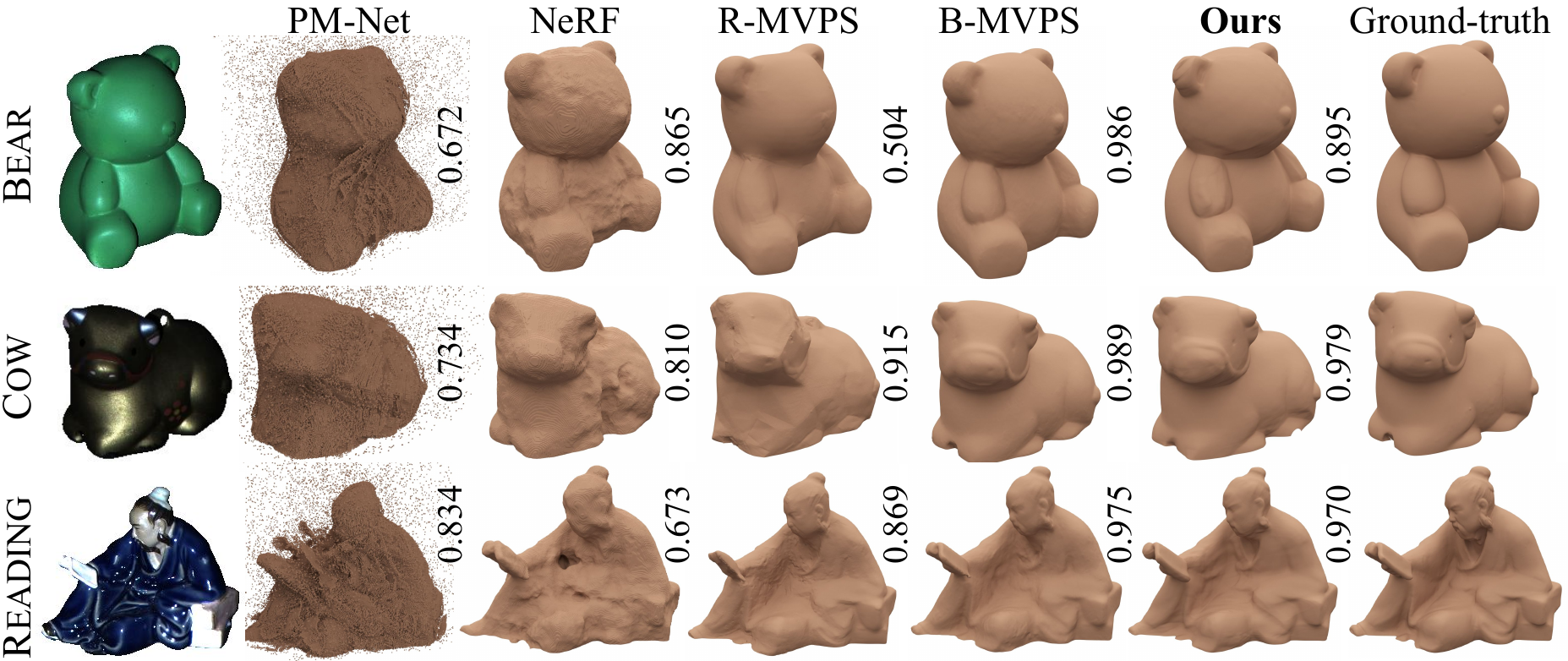}
    \caption{\footnotesize Comparison of the reconstruction quality with PM-Net\cite{wang2021patchmatchnet}, NeRF\cite{mildenhall2020nerf}, R-MVPS\cite{park2016robust}, and B-MVPS\cite{li2020multi} using $\mathcal{F}$-score metric. }\label{fig:visual_comparison}
\end{figure}

\begin{table*}[t]
\scriptsize
\centering
    \resizebox{1.0\textwidth}{!}
    {
    \begin{tabular}{c|c|c|c|c|c|c|c|>{\columncolor{red!20}}c}

    \rowcolor[gray]{0.90}
   \textbf{ Method Type $\rightarrow$} & \multicolumn{2}{c|}{Deep Multi-View Stereo} & \multicolumn{2}{c|}{View-Synthesis} &
    \multicolumn{3}{c|}{Photometric Stereo}  
    \\ 
    \hline
       \rowcolor[gray]{0.90}
       \textbf{Dataset}$\downarrow$ $|$ \textbf{Method} $\rightarrow$ & MVSNet \cite{yao2018mvsnet} &  PM-Net \cite{wang2021patchmatchnet} &  NeRF \cite{mildenhall2020nerf} & IDR \cite{yariv2020multiview} & Robust PS \cite{oh2013partial} & SDPS-Net \cite{chen2019self} & CNN-PS \cite{ikehata2018cnn} &  \textbf{Ours}  \\
        \hline
        BEAR     & 0.135 & 0.672 & 0.865 & 0.053 & 0.266 & 0.239 & 0.293 & \textbf{0.895} \\
        BUDDHA   & 0.147 & 0.799 & 0.713 & 0.150 & 0.367 & 0.298 & 0.363 & \textbf{0.922} \\
        COW      & 0.095 & 0.734 & 0.810 & 0.098 & 0.245 & 0.447 & 0.511 & \textbf{0.979} \\
        POT2     & 0.126 & 0.666 & 0.859 & 0.079 & 0.231 & 0.464 & 0.632 & \textbf{0.907} \\
        READING  & 0.115 & 0.834 & 0.673 & 0.073 & 0.242 & 0.188 & 0.508 & \textbf{0.970} \\
        \hline
        \textbf{AVERAGE} & 0.124  & 0.741 & 0.784 & 0.091 & 0.270 & 0.327 & 0.461 & \textbf{0.935} \\
    \hline
    \end{tabular}
    }
    \caption{\footnotesize $\mathcal{F}$-score comparison with standalone methods on DiLiGenT-MV dataset \cite{li2020multi}. The statistics show that our method achieves better results compared to the SOTA standalone multi-view and PS methods, thereby showing the advantage of using the complementary surface information in a MVPS setup.
    }
    \label{tab:standalone_numerical_comparison}
\end{table*}

\begin{figure*}[t]
    \centering
    \subfigure[\label{fig:tsdf_fusion}{$\mathcal{F}$-score comparison with TSDF Fusion}]
    {\includegraphics[height=0.17\textwidth, width=0.28\textwidth]{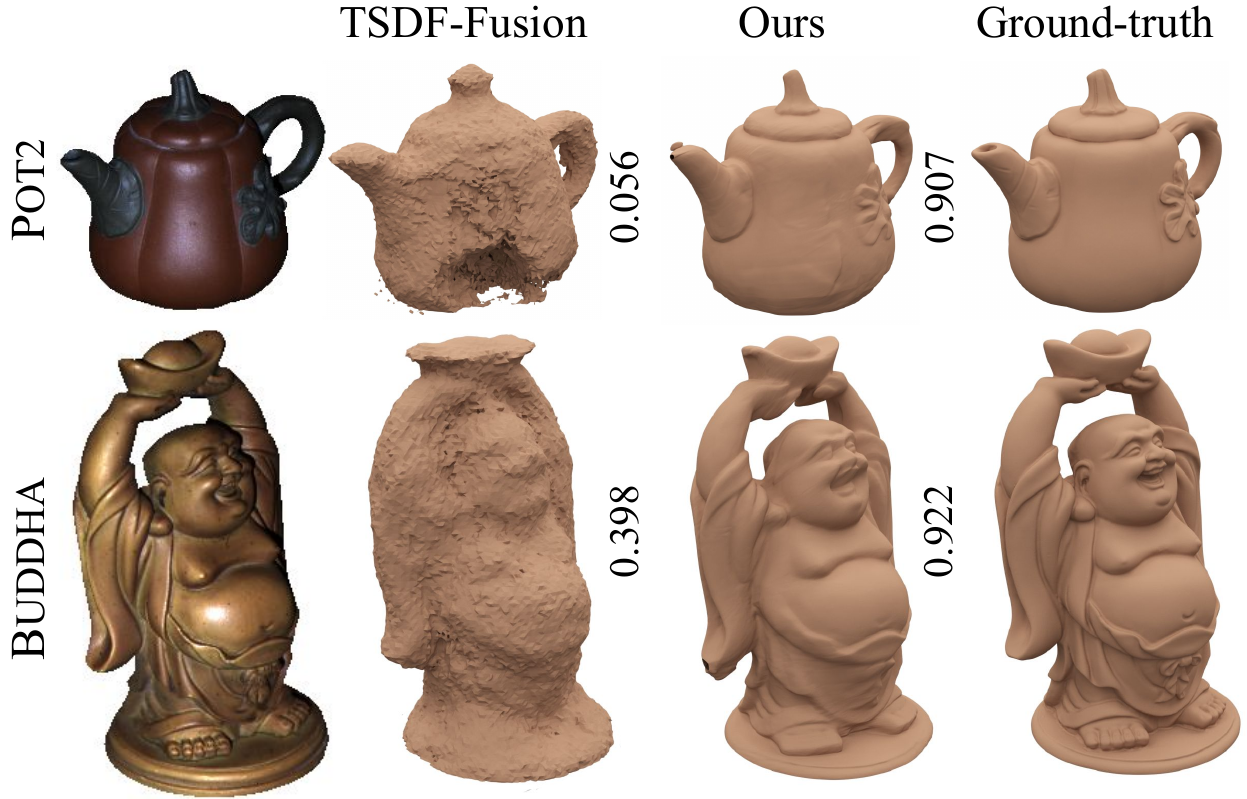}}
     \subfigure[\label{fig:noise_experiment}{Precision w.r.t. noise levels}]
    {\includegraphics[height=0.17\textwidth, , width=0.165\textwidth]{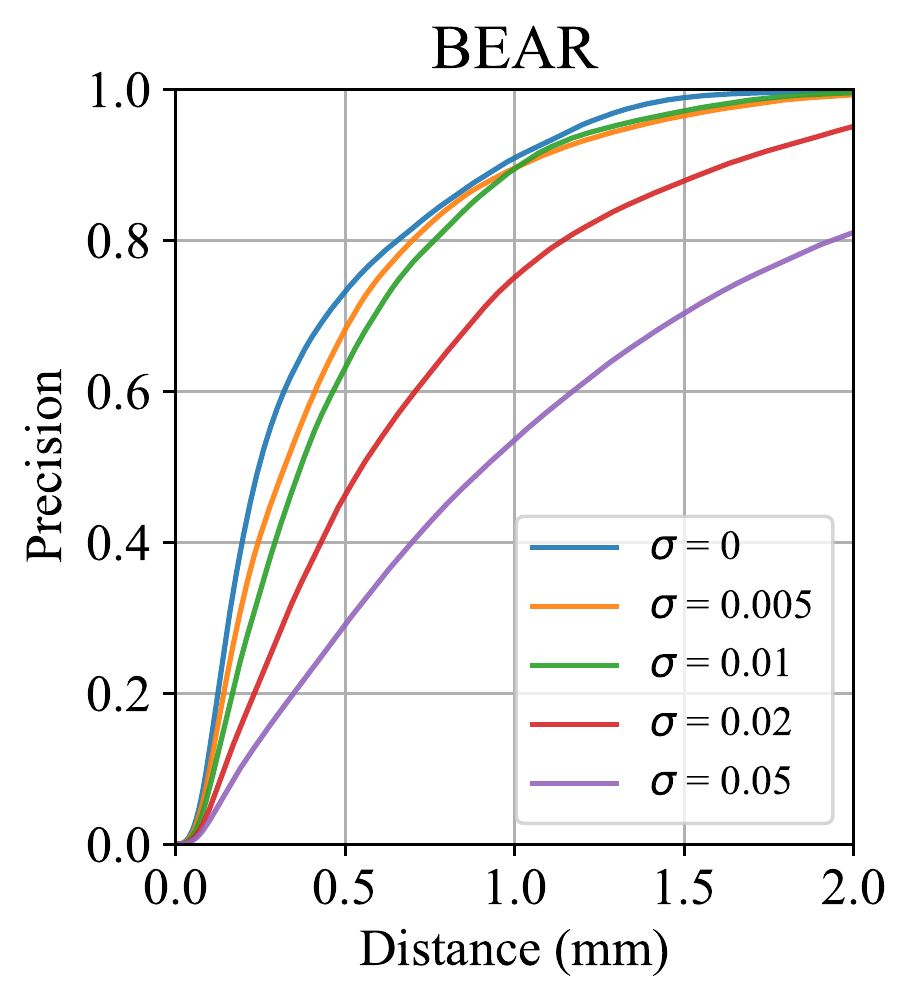}
        \includegraphics[height=0.17\textwidth, width=0.165\textwidth]{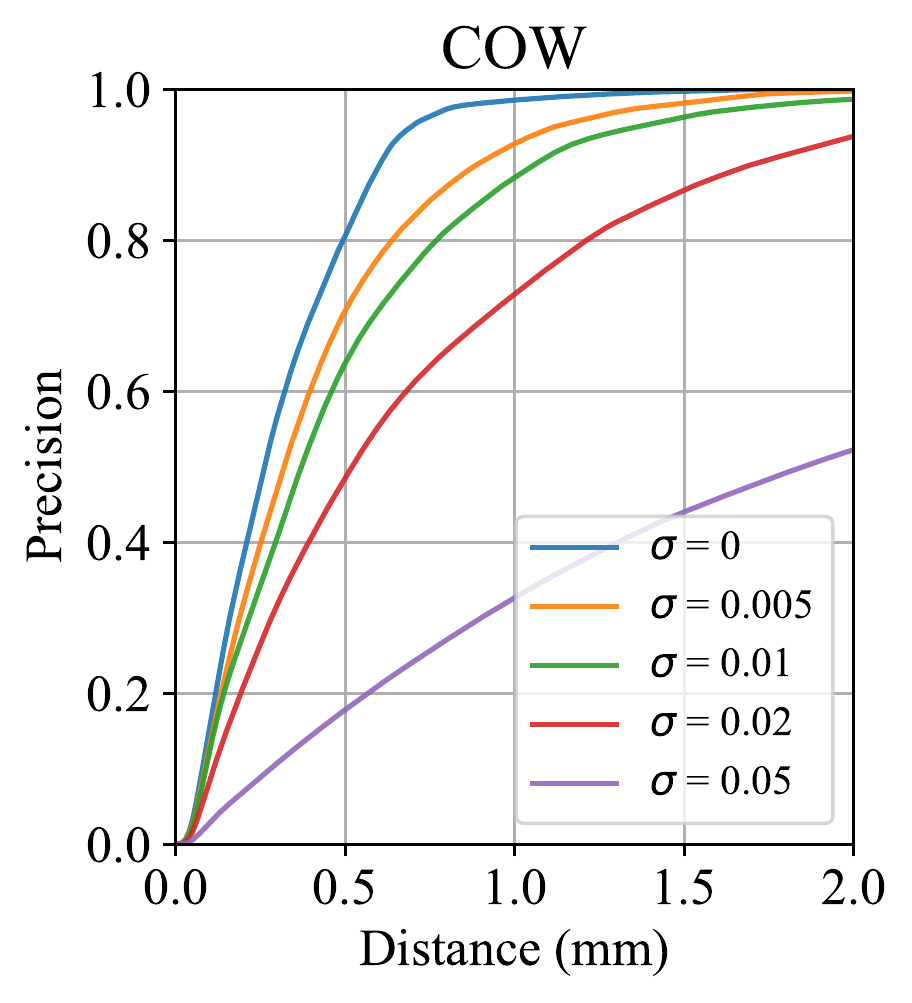}}
     ~\subfigure[\label{fig:profile_comparison}{Recovered surface profile}]
    {\includegraphics[height=0.17\textwidth, width=0.18\textwidth]{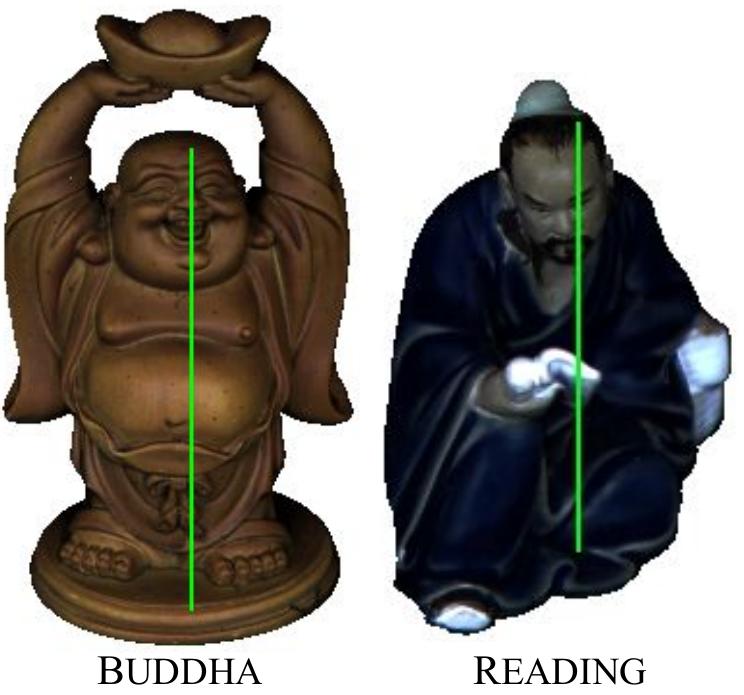}
    \includegraphics[height=0.17\textwidth, width=0.18\textwidth]{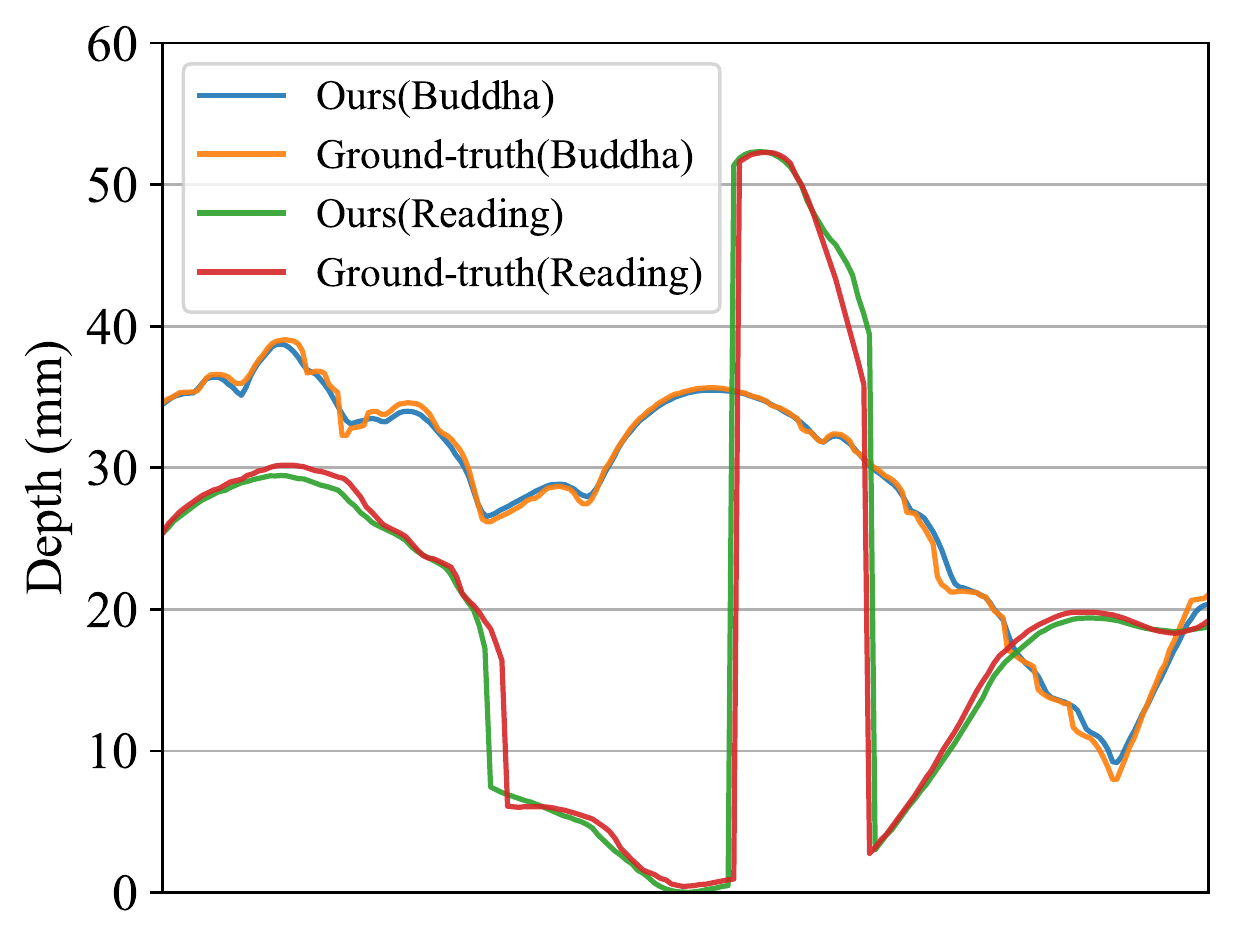}
    }
    \caption{\footnotesize (a) Comparison of the reconstruction quality with TSDF Fusion \cite{curless1996volumetric}, which is a standard method of choice for robust 3D fusion (outlier removal).  (b) Surface reconstruction accuracy for Bear and Cow objects under different noise levels. We report precision as a function of distance threshold ($\tau$) to show the fraction of accurately reconstructed points. (c) Estimated surface profile using our approach, showing how the recovered 3D shape follows the ground-truth surface profile curve when compared across the arbitrarily chosen geodesics. }
    \label{fig:ablation}
\end{figure*}

\noindent
\textit{\uline{(ii) Uncertainty-Aware Deep Multi-View Stereo.}}
The deep-MVS network is trained using DTU MVS dataset \cite{aanaes2016large}. It is trained for 8 epochs using Adam optimizer \cite{kingma2014adam} and learning rate of $0.001$. To predict depth with coarse-to-fine approach, we set number of stages $M=3$ 
for performing PatchMatch at different scales with $N_{iter}^3 = 2$, $N_{iter}^2 = 2$, $N_{iter}^1 = 1$ (higher $M$ indicates coarser scale). 
$\mathcal{D}_{f}=48$ depth hypotheses are used at initialization. For local perturbation
$N_{l}^3 = 16$, $N_{l}^2 = 8$, $N_{l}^1 = 8$, and  $N_d^3 =16$, $N_d^2 =8$, $N_d^1 =0$ at propagation steps. At last, depth refinement is performed at the original image resolution \cite{wang2021patchmatchnet}.

For testing, we use DiLiGenT-MV images with $N_s=2$ source images per reference image. We predict per-pixel depth and confidence measure $\rho_i$ for all views using the above parameters. We set  $\tau_{mvs} = 0.9$ to obtain $c_{i}^{mvs}$, where $c_{i}^{mvs}=1$ if $\rho_i > \tau_{mvs}$ and $c_{i}^{mvs}=0$ otherwise.

\noindent
\textit{\uline{(iii) Overall Loss Optimization.}}
We optimize for a zero-level set surface defined by highly confident estimates in $\mathcal{X}$ and $\mathcal{N}$ (see Sec. \S\ref{subsec:implicit_neural_representation}). For that, we first perform a precautionary multi-view consistency check to eliminate a few spurious 3d points. MLP with 8 layers is then used on remaining highly-confident estimates to learn suitable implicit neural shape representation. Here, each MLP layer contains 512 hidden units. A skip connection combines the input to the $4^{th}$ layer to speed up learning \cite{gropp2020implicit}. We set the parameters $\lambda_{n} = 10$ and $\lambda_{e} = 1$ for our loss function (Eq.\eqref{eq:level_set_optimization}). 
The distribution $\mathcal{P}$ for the expectation in Eq.\eqref{eq:level_set_optimization} is taken as average of (a) sum of Gaussian centered at points of $\mathcal{X}$ locally and (b) a uniform distribution globally. The standard deviation of Gaussian at a point is taken as the distance to the $50^{th}$ nearest neighbor. For optimization, we used Adam optimizer \cite{kingma2014adam} with a learning rate of $0.001$. We train the MLP for $10^5$ epochs by sampling $2^{14}$ elements from $\mathcal{X}, \mathcal{N}$  in each batch. We run the trained MLP on a volumetric grid of size $512^3$, which is then used by marching cubes algorithm \cite{lorensen1987marching} to extract the mesh corresponding to the zero level set of MLP based neural shape representation.

\subsection{Statistical Study}
\formattedparagraph{Evaluation Metrics.}
We evaluated our results using the Chamfer-$L_1$, precision, and $\mathcal{F}$-score metric as defined in \cite{knapitsch2017tanks}. In our evaluation, we used $\mathcal{F}$-score metric with $1 mm$ distance threshold $\tau$.  Next, we present our baseline results, which we have divided into three sub-categories.

\subsubsection{Baseline Comparisons}
\noindent
\formattedparagraph{(a) With standalone methods.}
Here, we compare our performance with methods that use either PS or MVS setup. Such an experiment helps us understand the benefit of using PS and MVS information together and how accurately we can reconstruct the shape with standalone methods. We used $\mathcal{F}$-score to compare our method's performance with SOTA PS, MVS, and view-synthesis methods. Table(\ref{tab:standalone_numerical_comparison}) provides the statistics for the same, indicating the clear advantage of our approach against the standalone methods.

\begin{table}[h]
    \centering
    \resizebox{\columnwidth}{!}
    {
    \begin{tabular}{c|c|c|c|>{\columncolor{red!20}}c|>{\columncolor{blue!15}}c}
    \hline
    \rowcolor[gray]{0.90}
       \textbf{Dataset}$\downarrow$ $|$ \textbf{Method} $\rightarrow$ & Kaya \etals \cite{kaya2021neural} & R-MVPS \cite{park2016robust} & B-MVPS \cite{li2020multi}   & \textbf{Ours} & Difference with \cite{li2020multi}  \\
        \hline
        BEAR     & 0.856 & 0.504 & \textbf{0.986} & \textcolor{RoyalBlue}{\textbf{0.895}}  & 0.091\\
        BUDDHA   & 0.690 & \textbf{0.935} & \textcolor{RoyalBlue}{\textbf{0.934}} & 0.922 & 0.012\\
        COW      & 0.844 & 0.915 & \textbf{0.989} & \textcolor{RoyalBlue}{\textbf{0.979}} & 0.010\\
        POT2     & 0.858 & 0.458 & \textbf{0.984} & \textcolor{RoyalBlue}{\textbf{0.907}} & 0.077 \\
        READING  & 0.720 & 0.869 & \textbf{0.975} & \textcolor{RoyalBlue}{\textbf{0.970}} & 0.005\\
        \hline
        \textbf{AVERAGE} & 0.794 & 0.736 & \textbf{0.974} & \textcolor{RoyalBlue}{\textbf{0.935}} & 0.039 \\
    \hline
    \end{tabular}
    }
    \caption{\footnotesize $\mathcal{F}$-score comparison with different MVPS methods on DiLiGenT-MV dataset \cite{li2020multi}. The statistics show that our method performs far better than R-MVPS and compares favorably with the SOTA B-MVPS \cite{li2020multi}. The point to note is that we can get results close to the SOTA with a simple and easy-to-implement method.}
    \label{tab:mvps_numerical_comparison}
\end{table}

\begin{figure*}[t]
    \centering
    \subfigure[\label{fig:texture_comparison}{Quality of textured mesh comparison.}]
    {\includegraphics[width=0.48\linewidth]{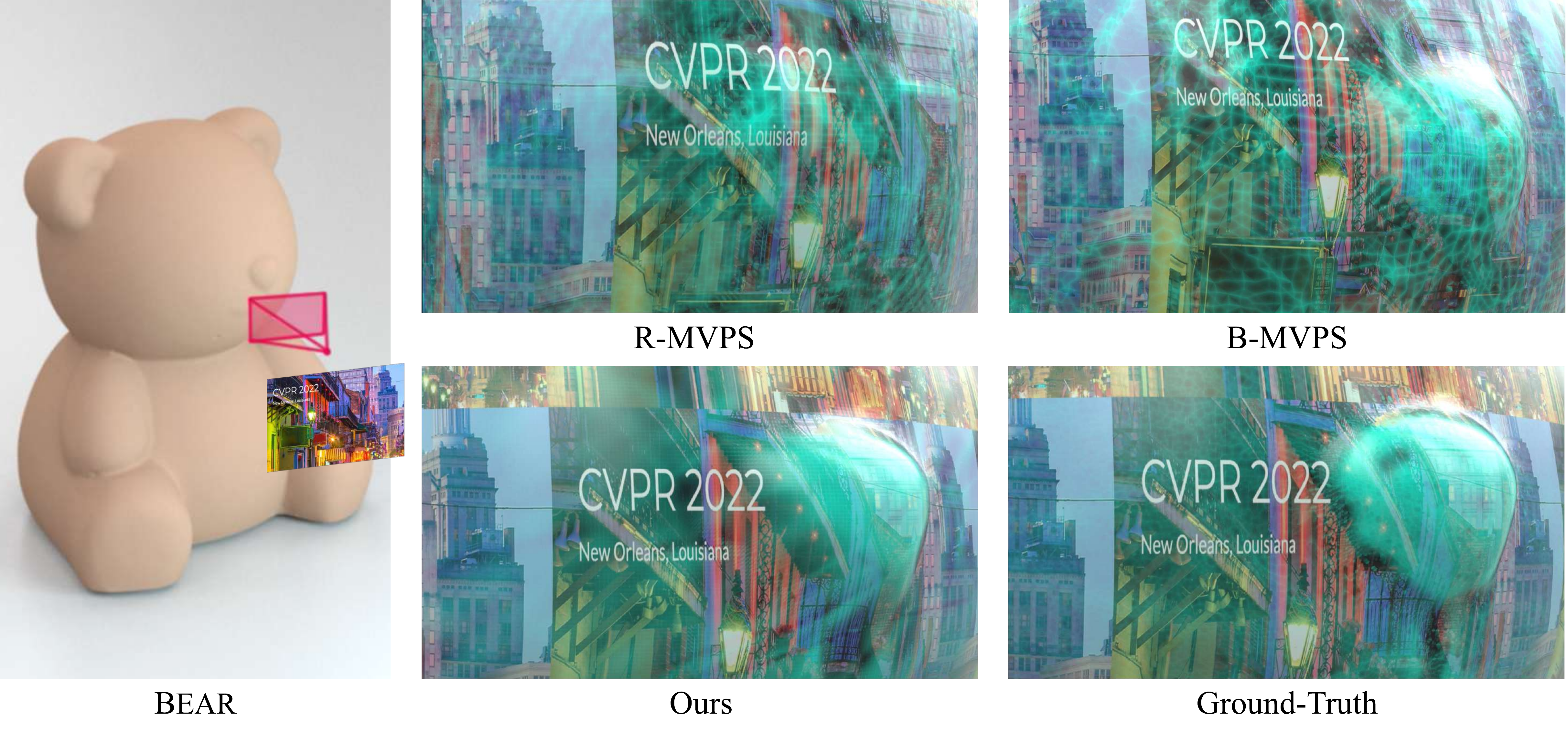}}
    ~~\subfigure[\label{fig:wireframe}{Wireframe visualization for 3D shape topology analysis.}]
    {\includegraphics[width=0.48\linewidth, height=0.222\linewidth]{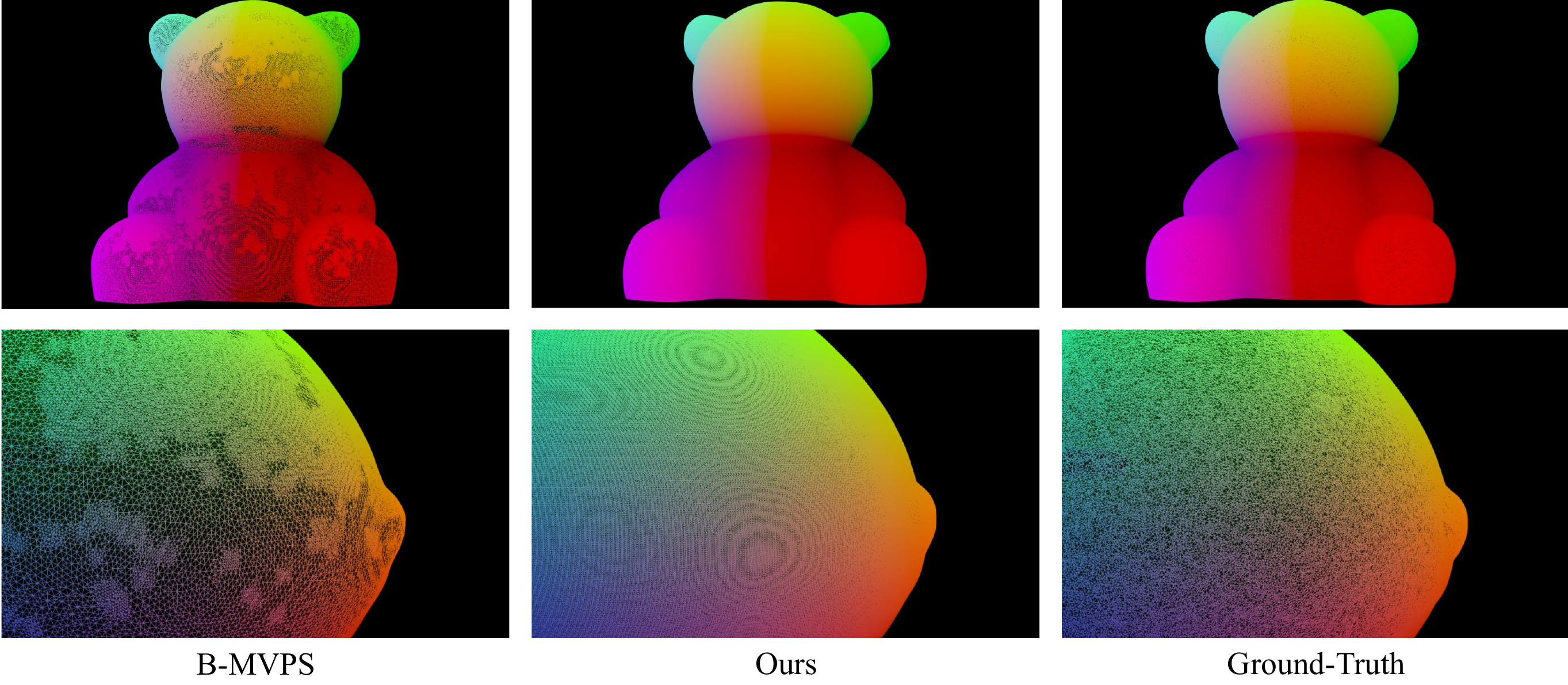}}
    \caption{\footnotesize (a) We transfer the CVPR'22 logo texture on the local region (around the nose) of the mesh recovered using SOTA MVPS methods. It can be observed that the texture pattern on our recovered mesh is closer to GT compared to B-MVPS \cite{li2020multi} and R-MVPS \cite{park2016robust} (notice the shift of the text). We want to emphasize that if the local topology is same, it
    must place the text at similar location as can be seen in ours result and GT.
    %
    (b) Colored Wireframe comparison with SOTA B-MVPS \cite{li2020multi}. Clearly, the distribution of geometric primitives on B-MVPS result is uneven and unbalanced compared to ours.}
    \label{fig:mesh_comparison}
\end{figure*}

\noindent
\formattedparagraph{(b) With MVPS methods.}
We compare our method with well-known MVPS methods, \ies, R-MVPS \cite{park2016robust}, B-MVPS \cite{li2020multi}, and Kaya \etals \cite{kaya2021neural}.  Table(\ref{tab:mvps_numerical_comparison}) provides $\mathcal{F}$-score comparison with these methods. Our method performs better than R-MVPS \cite{park2016robust}, Kaya \etals \cite{kaya2021neural} and compares favorably with B-MVPS \cite{li2020multi} with the minor difference in $\mathcal{F}$-score values \ies,  less than $10^{-1}$ in all object categories. Additionally, we want to emphasize again that B-MVPS \cite{li2020multi} relies on several carefully crafted explicit geometric steps and refinements that are complex and time-consuming, while our deep MVPS approach is easy to implement and realize.  

\smallskip
\noindent
\formattedparagraph{(c) With standard volumetric fusion method.}
For estimating the dense 3D surface, it is possible to combine the deep-MVS depth map and depth from PS surface normal map via widely used robust 3D fusion technique, \ies, TSDF fusion \cite{curless1996volumetric}.  To show that our approach is better at estimating the object's surface than TSDF fusion, we executed the TSDF fusion algorithm on our deep-MVS depth and depth from deep-PS output. The qualitative and quantitative comparison is presented in Fig.\ref{fig:tsdf_fusion}. The results clearly show that TSDF fusion provides undesirable output. On the other hand, our approach takes proper care of the surface estimates and provides much better 3D surface reconstruction.

\subsubsection{Ablation Study and Further Analysis}
\noindent
\formattedparagraph{(a) Effect of uncertainty modeling.}
To understand the impact of uncertainty modeling on our method's performance, we analyzed our results under three distinct settings of overall loss function  \ies, Eq.\eqref{eq:level_set_optimization}: $\textit{(i)}$~We remove both deep-MVS and deep-PS confidence variable, \ies, $c_{i}^{mvs}$, $c_{i}^{ps}$ from Eq.\eqref{eq:level_set_optimization} (w/o uncertainty modeling), $\textit{(ii)}$~We keep the deep-MVS confidence variable ($c_{i}^{mvs}$) and drop the deep-PS confidence variable ($c_{i}^{ps}$) from Eq.\eqref{eq:level_set_optimization} (w/o PS uncertainty modeling), and $\textit{(iii)}$~We keep the deep-PS confidence variable ($c_{i}^{ps}$) and drop the deep-MVS confidence variable ($c_{i}^{mvs}$) from Eq.\eqref{eq:level_set_optimization} (w/o MVS uncertainty modeling). Table(\ref{tab:ablation}) shows the Chamfer-$L_1$ metric results obtained under the three settings. The results indicate that incorporating uncertainty information to the loss helps handle the erroneous estimations, and therefore, results in better 3d reconstruction.

\begin{table}[h]
    \centering
    \resizebox{\columnwidth}{!}{
		\begin{tabular}{c|c|c|c|c|c|c}
			\hline
			\rowcolor[gray]{0.85}
			\textbf{Settings}$\downarrow$ $|$ \textbf{Dataset} $\rightarrow$ & BEAR & BUDDHA & COW  & POT2 & READING & \textbf{AVG}\\ \hline
			w/o Uncertainty Modeling & 0.468 &	0.485 &	0.365 &	0.557 &	0.380 &	0.451 \\ \hline
			w/o PS Uncertainty Modeling & 0.443 &	0.481 &	0.381 &	0.484 &	0.377 &	0.433  \\ \hline
			w/o MVS Uncertainty Modeling &  0.457 &	0.473 &	0.339 &	0.636 &	1.024 &	0.586  \\ \hline
			Ours (LCNet\cite{chen2019self} light) & 0.481 &	0.465 &	0.346 &	\textbf{0.481} &	0.381 &	0.431  \\ \hline
			\rowcolor{red!20} Ours & \textbf{0.415} &	\textbf{0.455} & \textbf{0.329} &	0.515 &	\textbf{0.355} & \textbf{0.414} \\ \hline
		\end{tabular}}
		\caption{\footnotesize Effect of uncertainty modeling and light calibration on the reconstruction quality of our approach. The results show Chamfer-$L_1$ metric (lower is better). The numbers confirm that utilizing the estimated uncertainty of both PS and MVS produces the best results. Further, our approach performs well without the exact light sources \ies, LCNet light sources \cite{chen2019self}.
		}
		\label{tab:ablation}
\end{table}

\noindent
\formattedparagraph{(b) Effect of light sources.} 
Here, we study the behavior of our approach under the uncalibrated-PS setting where GT light sources are not given. Instead, we used a pre-trained neural network to have an initial estimate of light sources. Precisely, we used LCNet \cite{chen2019self} model to have light source direction and intensity values. Table(\ref{tab:ablation}) shows Chamfer-$L_1$ metric obtained using LCNet predicted light sources information. Our method performs almost equivalently well, showing robustness to small errors in the light calibration.

\smallskip
\noindent
\formattedparagraph{(c) Effect of noise.}
We consider the MVPS acquisition setup in our work, where imaging noise is inevitable. So, we analyze the behavior of our approach under imaging noise. To that end, we add zero-mean Gaussian noise to images with different standard deviations ($\sigma$'s). Fig.\ref{fig:noise_experiment} shows the precision curve of the recovered surface as a function of the distance threshold $\tau$. Precisely, it measures the fraction of points that are reconstructed accurately. The plots show that increasing the noise level degrades the performance. Further, we infer that behavior among subjects varies as signal-to-noise ratio changes. We can observe that our method is robust, and the performance drop is not random.

\smallskip
\noindent
\formattedparagraph{(d) Quality of reconstructed surface geometry.} 
To perform this experiment, we first analyzed the surface profile of our reconstructed shape across randomly chosen geodesics on the object. Fig.\ref{fig:profile_comparison} shows that our recovered surface profile closely follows GT. Next, we performed a local surface analysis of recovered mesh.  Although evaluation of recovered shape based on global performance metric is already discussed, it may not reflect the true picture of surface topology since the actual distribution of mesh on GT shape is not known a priori. So, as a second experiment, we transferred texture on a local mesh topology and qualitatively compared the results. Fig.\ref{fig:texture_comparison} shows texture transfer results on the mesh recovered using different methods, and clearly, our textured mesh reflects fine text details and appears close to GT. Further, we analyzed the colored Wireframe of the recovered shape compared to SOTA B-MVPS \cite{li2020multi}. Fig.\ref{fig:wireframe} Wireframe model shows that the distribution of geometric primitives in our recovered shape is smooth, regular, and close to GT, whereas B-MVPS \cite{li2020multi} has an unbalanced distribution of geometric primitives.

\smallskip
\formattedparagraph{Limitations.}
This work assumed light sources and camera calibration information are given as input. Further, we tested our method on \cite{li2020multi} dataset, which is generally composed of isotropic material objects, and it will be interesting to investigate our method's performance on anisotropic material objects. Nevertheless, for now, we are limited by the availability of open-source MVPS datasets.

\section{Conclusion}
This paper explores the field of MVPS using the concepts from deep learning, geometry, and uncertainty. Unlike existing MVPS methods, which treat 3D shape reconstruction as point estimation and geometric optimization problems, we propose learning the fidelity of surface estimation and recovering the shape based on the implicit neural shape representation. Without using complex geometric steps, we observed that our simple neural network based approach could provide results comparable to the best available algorithm. Thus, we believe our work will enable broader use of the MVPS approach in precise 3D data acquisition.

\formattedparagraph{Acknowledgement.} {\footnotesize {This work was supported by Focused Research Award from Google, ETH Z\"urich Foundation (CVL, ETH 2019-HE-318, 2019-HE-323, 2020-FS-351, 2020-HS-411).}}

\nocite{*}
{\small
\balance
\bibliographystyle{ieee_fullname}
\bibliography{ReviewTemplate}
}


\end{document}